\documentclass[letterpaper]{article} 
\usepackage[draft]{aaai25}  
\usepackage{times}  
\usepackage{helvet}  
\usepackage{courier}  
\usepackage[hyphens]{url}  
\usepackage{graphicx} 
\usepackage{natbib}  
\usepackage{caption} 
\usepackage{algorithm}
\usepackage{algorithmic}

\usepackage{newfloat}
\usepackage{listings}

\usepackage{amsmath} 
\usepackage{multirow}
\usepackage{tabularx}
\usepackage{booktabs}
\usepackage{amssymb}
\usepackage{hhline}
\usepackage{pifont}
\usepackage{utfsym}
\usepackage{multicol}
\usepackage{tabularx} 
\usepackage{colortbl}

\DeclareCaptionStyle{ruled}{labelfont=normalfont,labelsep=colon,strut=off} 
\floatstyle{ruled}
\newfloat{listing}{tb}{lst}{}
\floatname{listing}{Listing}
\title{CrossDiff: Diffusion Probabilistic Model With Cross-conditional Encoder-Decoder for Crack Segmentation}
\author{
    Xianglong Shi\textsuperscript{\rm 1}, Yunhan Jiang\textsuperscript{\rm 1}, Xiaoheng Jiang\textsuperscript{\rm 1}\thanks{Corresponding Author}, Mingling Xu\textsuperscript{\rm 1}, Yang Liu\textsuperscript{\rm 2}\\
}
\affiliations{
    \textsuperscript{\rm 1}Zhengzhou University\\
    \textsuperscript{\rm 2}King's College London\\

    csxlshi@163.com, yhjiang@stu.zzu.edu.cn, \{jiangxiaoheng,  iexumingliang\}@zzu.edu.cn, yang.9.liu@kcl.ac.uk
}

\usepackage{bibentry}

\begin{document}

\maketitle

\begin{abstract}
Crack Segmentation in industrial concrete surfaces is a challenging task because cracks usually exhibit intricate morphology with slender appearances. Traditional segmentation methods often struggle to accurately locate such cracks, leading to inefficiencies in maintenance and repair processes. In this paper, we propose a novel diffusion-based model with a cross-conditional encoder-decoder, named CrossDiff, which is the first to introduce the diffusion probabilistic model for the crack segmentation task. Specifically, CrossDiff integrates a cross-encoder and a cross-decoder into the diffusion model to constitute a cross-shaped diffusion model structure. The cross-encoder enhances the ability to retain crack details and the cross-decoder helps extract the semantic features of cracks. As a result, CrossDiff can better handle slender cracks. Extensive experiments were conducted on five challenging crack datasets including CFD, CrackTree200, DeepCrack, GAPs384, and Rissbilder. The results demonstrate that the proposed CrossDiff model achieves impressive performance, outperforming other state-of-the-art methods by 8.0\% in terms of both Dice score and IoU. The code will be open-source soon.
\end{abstract}

%
\section{Introduction}
\label{sec:intro}
Industrial concrete surfaces, such as roads and pavements~\cite{koch2015review,konig2022s,kheradmandi2022critical,du2023modeling}, are susceptible to the formation of cracks over time due to various factors including traffic load~\cite{nguyen2023deep}, temperature fluctuations~\cite{kheradmandi2022critical}, and environmental conditions~\cite{munawar2021image}. These cracks not only compromise the structural integrity of the surfaces but also pose safety hazards to vehicles and pedestrians. Effective maintenance and repair strategies necessitate accurate segmentation of these cracks to identify their extent and severity~\cite{huthwohl2018detecting,choi2019sddnet,panella2022semantic}. However, segmentation of cracks remains a challenging task due to their intricate morphology and the presence of noise and artifacts in the concrete surface images.

\begin{figure*}
    \centering
    \includegraphics[width=0.85\linewidth]{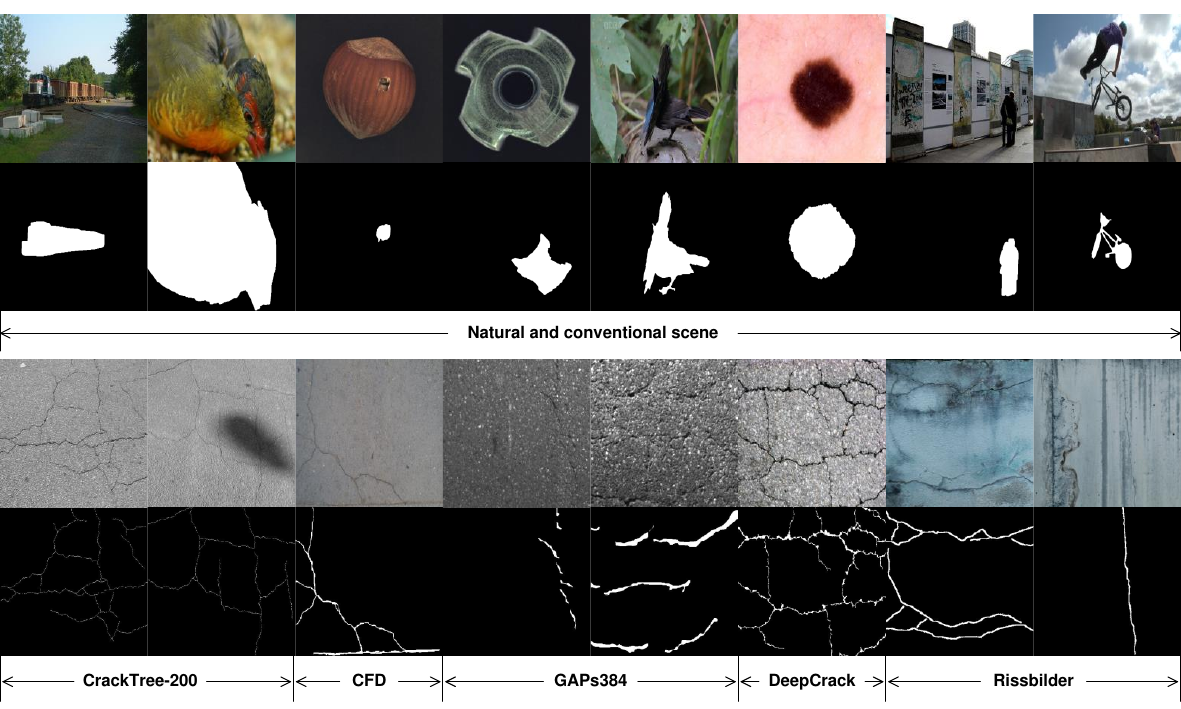}
    \caption{Comparison between natural and conventional scene images with industrial crack images. The top row showcases images from mainstream semantic segmentation datasets, including Pascal VOC~\cite{everingham2015pascal}, MVTecAD~\cite{bergmann2019mvtec}, SegTrack-v2~\cite{li2013video} and ISIC~\cite{gutman2016skin}, with corresponding ground truth images in the second row. The third row displays samples from CFD~\cite{shi2016automatic},  CrackTree-200~\cite{zou2012cracktree}, DeepCrack~\cite{liu2019deepcrack}, GAPs384~\cite{eisenbach2017get}, and Rissbilder~\cite{pak2021crack} with their ground truth images below.}
    \label{fig:example}
\end{figure*}

In addition to their complex morphology, as shown in Figure~\ref{fig:example}, long-narrow cracks often manifest low contrast or irregular intensity profiles, further complicating their segmentation. Variations in illumination, imaging conditions, or material properties can obscure these cracks, making their detection and delineation challenging~\cite{khan2023image}. The presence of noise and artifacts in real-world imaging scenarios further compounds the segmentation task by introducing spurious features that may interfere with accurate crack identification.

Deep learning techniques, particularly Convolutional Neural Networks (CNNs)~\cite{zhou2018unet++,he2017mask,chen2018encoder} and attention mechanisms (as seen in Transformer models~\cite{cheng2021mask2former,xie2021segformer,kirillov2023segment}), have demonstrated remarkable success in various image processing tasks, especially in segmentation~\cite{xie2021segformer,cheng2021maskformer,cheng2021mask2former}. However, applying these techniques to the segmentation of slender cracks in concrete surfaces still presents unique challenges. For instance, CNN-based methods~\cite{choi2019sddnet,pang2022dcsnet,du2023modeling} are inherently limited in capturing fine-grained details, especially in regions with subtle variations in texture and color. Additionally, slender cracks often exhibit irregular shapes and low contrast~\cite{tabernik2023automated}, making them difficult to distinguish from background clutter and noise. While attention mechanisms, such as those employed in Transformer models~\cite{zhou2023hybrid,ding2023crack,guo2023pavement}, can help prioritize relevant information~\cite{vaswani2017attention}, they still struggle to effectively capture the spatial relationships between pixels in long and narrow structures.

In this study, we introduce CrossDiff, a novel Diffusion Probabilistic Model (DPM)~\cite{ho2020denoising,nichol2021improved} architecture with cross-conditional encoding and decoding, for slender crack segmentation. 
By integrating cross-conditional information exchange and probabilistic modeling, CrossDiff aims to overcome the inherent challenges associated with detecting and delineating slender cracks across various imaging modalities. Our approach leverages the strengths of deep learning to learn informative representations from data while incorporating probabilistic modeling to capture uncertainty and enhance segmentation accuracy. Through comprehensive experimental evaluations on diverse datasets, including CFD~\cite{shi2016automatic},  CrackTree-200~\cite{zou2012cracktree}, DeepCrack~\cite{liu2019deepcrack}, GAPs384~\cite{eisenbach2017get}, and Rissbilder~\cite{pak2021crack}, we demonstrate the efficacy and generalization capabilities of CrossDiff in addressing the challenging task of slender crack segmentation.

In summary, our paper makes the following contributions:
\begin{itemize}
    \item The first to introduce Diffusion Probabilistic Model-based framework for crack detection and prove its efficiency toward slender crack segmentation, as evidenced by the formulaic validation and experimental data.
    \item We propose a novel Cross-Conditional DPM architecture, CrossDiff, which takes advantage of a Cross Encoder and a Cross Decoder to jointly enhance the ability to extract crack features.
    \item Our proposed model demonstrates superior segmentation performance compared to other state-of-the-art methods on all five challenging crack datasets, achieving notable improvements in terms of Intersection over Union (IoU) and Dice score.
    
\end{itemize}

\section{Related Work}
\subsection{Discriminant Segmentation}
After the introduction of pixel-level asphalt crack detection utilizing CNN models by Zhang~\cite{zhang2017automated}, several more precise approaches have emerged for analyzing pavement damage employing deep neural networks~\cite{zou2018deepcrack,yang2019feature,fei2019pixel}. For instance, Liu ~\cite{maeda2018road} introduced a network for pyramid features aggregation along with a post-processing scheme employing Conditional Random Fields (CRFs) to enhance crack segmentation. Zou ~\cite{zou2018deepcrack} proposed a multi-stage fusion approach built upon the SegNet encoder-decoder architecture to refine crack segmentation. Yang ~\cite{yang2019feature} presented a hierarchical boosting network combined with a feature pyramid for pavement crack detection, effectively integrating contextual information into low-level features in a feature-pyramid manner. Fei ~\cite{fei2019pixel} introduced the CrackNet-V model, which incorporates stacked 3 × 3 convolutional layers and a 15 × 15 convolution kernel to achieve deep abstraction and high performance in crack segmentation. 

Then CrackFormer~\cite{9711107} employs a Transformer network architecture featuring multi-head attention mechanisms, positional encoding, and an encoder-decoder framework to effectively capture long-range dependencies and spatial relationships within images, enabling precise fine-grained crack detection across various surfaces and conditions. Tabernik~\cite{tabernik2023automated} proposes a novel approach that combines per-pixel segmentation and per-image classification. Their model leverages an encoder-decoder architecture to achieve accurate segmentation results, while also maintaining low computational costs for real-time applications. Despite the promising results achieved by these segmentation-based crack detection methods, they often fall short in attaining satisfactory pixel-level segmentation precision, resulting in blurred and coarse segmentation outcomes.
\subsection{Generative Segmentation}
Generative segmentation, a burgeoning field in image analysis, has witnessed remarkable progress in recent years. Various approaches, including classical methods such as Markov Random Fields (MRFs) and Conditional Random Fields (CRFs), which model pixel dependencies, and deep generative models like Generative Adversarial Networks (GANs) and Variational Autoencoders (VAEs), have been explored for generative segmentation tasks. GANs produce realistic images through adversarial training, while VAEs learn latent representations and reconstruct images. Recently, the Diffusion Probabilistic Model (DPM) has emerged as a prominent topic in computer vision, attracting significant attention within the research community. Its applications in image generation, such as DALLE3~\cite{betker2023improving}, Imagen~\cite{saharia2022photorealistic}, and Stable Diffusion~\cite{rombach2022high}, have showcased remarkable generation capabilities, sparking extensive discussions. Moreover, recent studies have demonstrated its utility in image segmentation tasks, such as DMOIISE~\cite{wolleb2022diffusion}, SegDiff~\cite{amit2021segdiff}, MedSegDiff~\cite{wu2024medsegdiff} and MedSeg-\\Diff-v2~\cite{wu2024medsegdiff}.

\section{Method}
In this section, we present the methodology employed in our proposed approach, CrossDiff, for the segmentation of slender cracks in industrial concrete surfaces. We present the foundational component of our method, the Diffusion Probabilistic Model (DPM), which serves as the backbone for capturing spatial dependencies and promoting coherence in the segmentation results. Following this, we present the Cross-conditional network, a novel architecture designed to enhance feature representation and capture fine-grained details, particularly in regions with subtle variations. In final, we present the training process and its impact on the overall architecture of our model, elucidating how these components synergistically contribute to achieving superior performance in crack segmentation tasks.

\begin{figure*}
    \centering
    \includegraphics[width=0.85\textwidth]{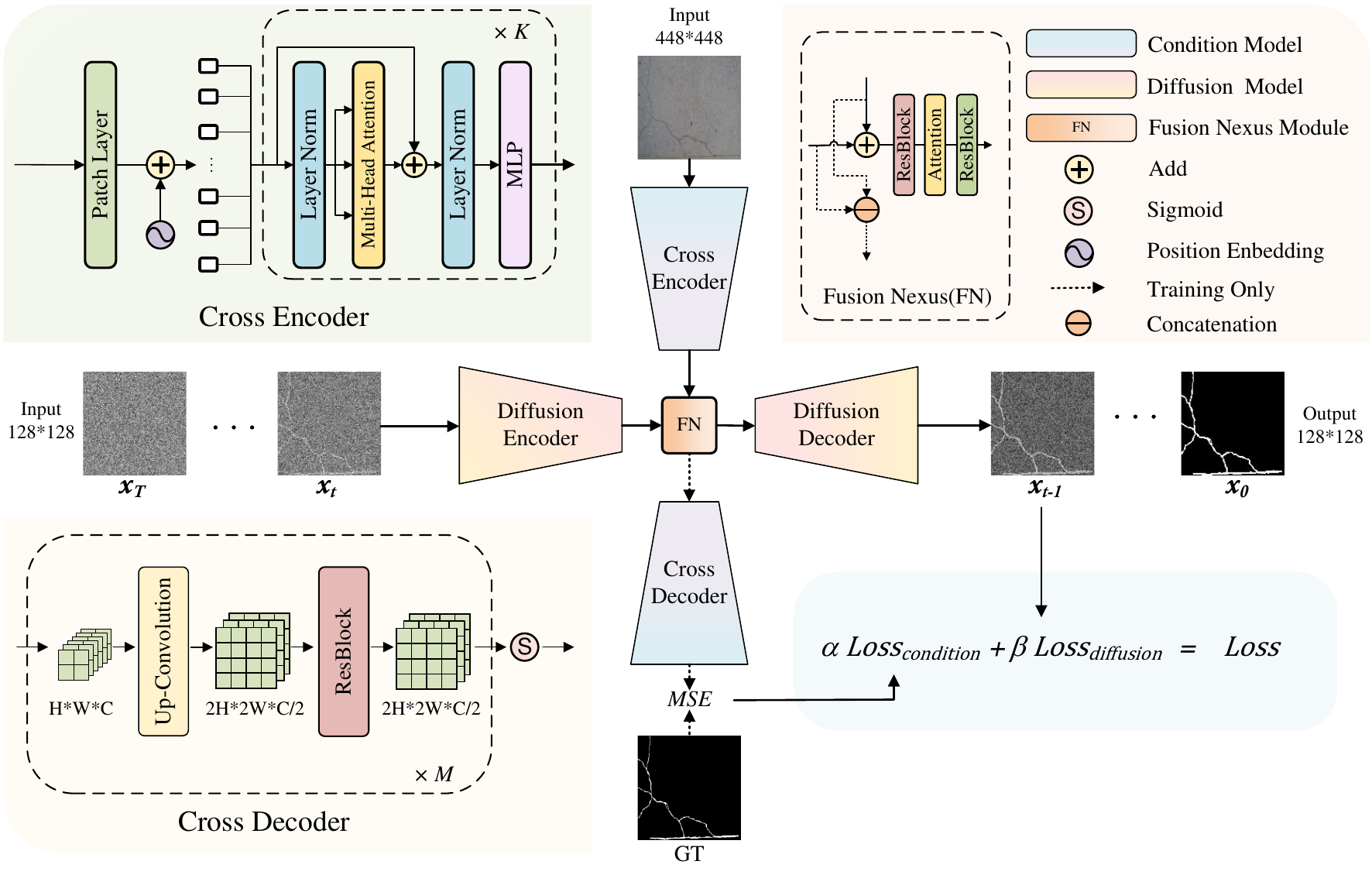}
    \caption{An illustration of CrossDiff. CrossDiff consists of Diffusion backbone and cross-conditional encoder-decoder, including Cross Encoder, Fusion Nexus, and Cross Decoder, which jointly constitute the cross-shaped diffusion model structure.}
    \label{fig:pipeline}
\end{figure*}

\subsection{Towards Generative Slender Segmentation}
Our model architecture draws inspiration from the diffusion model discussed in ~\cite{ho2020denoising} and ~\cite{nichol2021improved}, which constitutes a generative framework comprising a forward diffusion stage and a reverse diffusion stage. In the forward process, Gaussian noise is gradually incorporated into the segmentation label \( x_0 \) over a series of iterations \( T \). Conversely, the reverse process employs a neural network to reconstruct the original data by undoing the noise addition, formulated as:
\begin{equation}
p(x_{0:T-1} | x_T) = \prod_{t=1}^{T} p(x_{t-1} | x_t), 
\end{equation}
where \( x_{0:T-1} \) represents the sequence of latent variables, and \( x_T \) is the final noisy image. To ensure symmetry with the forward process, the reverse diffusion recovers the noise image step by step, ultimately yielding a clear segmentation.

In our implementation, we adopt a UNet architecture as the neural network for learning, as depicted in Figure~\ref{fig:pipeline} (Diffusion Encoder and Diffusion Decoder). To facilitate segmentation, we condition the step estimation function \( \epsilon \) by incorporating raw image priors:
\begin{equation}
\epsilon(x_t, I, t) = D((E_{I_t} + E_{x_t}, t), t), 
\end{equation}
where \( E_{I_t} \) denotes the conditional feature embedding derived from the raw image by Cross Encoder, while \( E_{x_t} \) represents the segmentation map feature embedding at the current step. These embeddings are combined and processed by a UNet decoder \( D \) for reconstruction, with the step-index \( t \) integrated into the process via a shared learned look-up table, following~\cite{ho2020denoising}.

Diffusion segmentation typically relies on the construction of a graph or network, where nodes represent pixels and edges represent connections between pixels. The weights of connections are usually determined based on the similarity between pixels. Let \( W_{ij} \) represent the weight of the connection between nodes \( i \) and \( j \).
For slender objects, we can assume that pixels within the object have high similarity with each other, as they may share similar color, texture, or shape features. We start with an image \( I \) containing a particularly slender object \( O \) and background \( B \). Our goal is to segment the object \( O \) from the background \( B \). Therefore, for pixels \( i \) and \( j \) within the object \( O \), the connection weight \( W_{ij} \) might be relatively large. Conversely, connections between the object \( O \) and the background \( B \) may have smaller weights.

Diffusion segmentation involves an information propagation process, where pixel labels (indicating whether they belong to object \( O \) or background \( B \)) affect the labels of their neighboring pixels. This process can be represented as:
\begin{equation}
L_i^{(t+1)} = \sum_{j \in N(i)} W_{ij} \cdot L_j^{(t)}, 
\end{equation}
where \( L_i^{(t)} \) represents the label of pixel \( i \) at iteration \( t \), and \( N(i) \) denotes the set of neighboring pixels of pixel \( i \).
Since pixels within the slender object have high similarity, with larger connection weights, they are more likely to influence each other during the information propagation process, leading to consistent labels within the object. On the other hand, for the background \( B \), where connection weights are smaller, the influence of information propagation may be comparatively weaker.

Thus, through the information propagation process, diffusion segmentation can more easily and stably segment the slender object from the background. This is why diffusion segmentation performs better when dealing with particularly slender objects, leveraging the similarity between pixels during the information propagation process to form consistent regions within the object.

\subsection{Cross-Conditional Encoder-Decoder}
\subsubsection{Cross Encoder}
The Cross Encoder, as shown in Figure~\ref{fig:pipeline}, inspired by the Vision Transformer (ViT)~\cite{dosovitskiy2020image,kirillov2023segment} backbone, consists of patch embedding, positional encoding, Transformer blocks, and a neck module. Patch embedding divides the input image into patches and projects them into a lower-dimensional space. Absolute positional embeddings are added to provide spatial information. Transformer blocks, comprising self-attention and MLP layers, capture global and local dependencies. Finally, the neck module refines the feature representations. The features extracted by this architecture effectively condition the step estimation function in diffusion probabilistic models, enabling precise segmentation of slender cracks in industrial concrete surfaces.
\subsubsection{Cross Decoder}
The Decoder module (Figure~\ref{fig:pipeline}), is pivotal in reconstructing segmented images from the conditioned features provided by the Conditional Encoder and Diffusion Encoder. It consists of a series of transposed convolutional layers, interleaved with residual blocks, to ensure efficient feature extraction and high-fidelity image reconstruction. Each ResidualBlock within the decoder comprises two transposed convolutional layers, followed by rectified linear unit (ReLU) activation functions and residual connections, facilitating effective learning of residual mappings. The decoder architecture encompasses six transposed convolutional layers, progressively increasing the spatial resolution by a factor of two through each layer, effectively capturing fine-grained details in the reconstruction process. After each transposed convolutional layer, a corresponding ResidualBlock is employed to refine the reconstructed features further. The final layer of the decoder comprises a transposed convolutional operation followed by a sigmoid activation function, yielding the ultimate segmentation masks. This comprehensive architecture enables the model to decode conditioned features accurately into a segmentation map.
\subsubsection{Fusion Nexus Module}
The middle layer acts as a pivotal bridge between the conditioned Decoder output and the Diffusion Encoder backbone, leveraging the current diffusion step information to intricately balance and fuse features from both sources. This layer consists of residual and attention blocks which dynamically adjust the importance of features according to the diffusion step to ensure adaptive information fusion. The Residual Blocks incorporate time embedding to capture temporal dynamics, while the Attention Block enables context-aware feature adjustment. Through residual connections and attentive processing, the middle layer refines the fused features, optimizing their utilization for segmentation tasks. 

\subsection{Optimization Objectives}
To optimize the diffusion encoder and cross encoder, the training loss is computed based on a modified version of Equation in ~\cite{ho2020denoising}, represented as follows:
\begin{equation}
    \resizebox{0.9\hsize}{!}{$
    \textit{Loss} =  \alpha {E}_{x_0, \epsilon, x_e, t} \left[ \left\| \epsilon - \epsilon_\theta \left( \sqrt{\bar{\alpha}_t} x_0 + \sqrt{1 - \bar{\alpha}_t} x_e, I_i, t \right) \right\| _2^2 \right]  + \beta \left( {x}_{d,t} - x_0 \right)^2.
$}
\end{equation}
during training, where \(x_0\) represents the segmentation of the input image \(I_i\) and \(x_{d,t}\) represents the segmentation output from cross decoder in each iteration, so the loss is computed by setting \(x_0 = S\), where \(S\) denotes the ground truth segmentation mask. 
In each iteration, a random pair of raw image \( I_i \) and its corresponding segmentation label \( S_i \) are sampled for training. The iteration number is sampled from a uniform distribution, and \( \epsilon \) is sampled from a Gaussian distribution.
Furthermore, the design of loss function allows the loss of Cross Decoder to impact both Diffusion Encoder and Cross Encoder. Enhancing the feature extraction capabilities of Diffusion Encoder for \(x_0\) in each step is effective for the final prediction.

\section{Experiments}
In this section, we present our experimental evaluation of the proposed method for the surface crack segmentation task. We compare our method with several related state-of-the-art approaches with publicly available code. 

\subsection{Dataset}
To validate the effectiveness of our model, we train and evaluate our network on one of the largest datasets proposed
by Li~\cite{li2021sccdnet} which they compiled from previously published
datasets CFD~\cite{shi2016automatic}, CRACK500~\cite{dorafshan2018sdnet2018}, CrackTree200~\cite{zou2012cracktree}, DeepCrack~\cite{liu2019deepcrack}, GAPs384~\cite{eisenbach2017get}, Rissbilder~\cite{pak2021crack},  Non-crack~\cite{dorafshan2018sdnet2018}, containing crack of road pavements, asphalt and concrete structures and walls. Datasets totally contain 7169 with manually annotated labels in a resolution of 448 × 448 pixels. It includes crack images from different environments, with varying sizes, shapes, shooting distances, and covering common crack features to the greatest extent possible.

\subsection{Evaluation Metrics}

The performance of our model for segmentation is assessed using two commonly employed evaluation metrics: Intersection over Union (IoU) and Dice coefficient. IoU measures the degree of overlap between the predicted segmentation mask and the ground truth mask. It is calculated as the ratio of the intersection area between the predicted and ground truth masks to their union. A higher IoU score indicates better alignment between the predicted and ground truth anomalies.The Dice coefficient, also known as the F1 score, quantifies the similarity between the predicted and ground truth segmentation masks. It is computed as twice the intersection area divided by the sum of areas of the predicted and ground truth masks. Like IoU, a higher Dice coefficient signifies better segmentation accuracy.


\subsection{Implementation Details}

 The proposed architecture is implemented in PyTorch framework and trained/tested on a 4090 GPU with 24GB of memory. All images are uniformly resized to the dimension of 128×128 pixels for Diffusion Encoder and 448×448 pixels for Cross Encoder. The networks are trained in an end-to-end manner using AdamW~\cite{loshchilov2017decoupled} optimizer. CrossDiff is trained with 12 batch size. The learning rate is initially set to 1e-4. The Cross Decoder is only utilized during training. All models are set 5 times of ensemble in the inference. We use STAPLE~\cite{warfield2004simultaneous} algorithm to fuse the different samples. 
\subsection{Main Results}

\begin{table*}
\renewcommand\arraystretch{1.0}
\begin{center}
\resizebox{\textwidth}{!}
{%
\begin{tabular}
    {>{\centering\arraybackslash}m{3cm}>{\centering\arraybackslash}m{1cm}>{\centering\arraybackslash}m{1cm}>{\centering\arraybackslash}m{1cm}>{\centering\arraybackslash}m{1cm}>{\centering\arraybackslash}m{1cm}>{\centering\arraybackslash}m{1cm}>{\centering\arraybackslash}m{1cm}>{\centering\arraybackslash}m{1cm}>{\centering\arraybackslash}m{1cm}>{\centering\arraybackslash}m{1cm}}
    \toprule
    \multirow{2}{*}{\centering Method} & \multicolumn{2}{c}{CFD} & \multicolumn{2}{c}{CrackTree200} & \multicolumn{2}{c}{DeepCrack} & \multicolumn{2}{c}{GAPs384} & \multicolumn{2}{c}{Rissbilder} \\ 
    \cmidrule(lr){2-3} \cmidrule(lr){4-5} \cmidrule(lr){6-7} \cmidrule(lr){8-9} \cmidrule(lr){10-11}
                         & Dice & IoU & Dice & IoU & Dice & IoU & Dice &IoU & Dice & IoU             \\
    \midrule
    DeepLabv3+   &  73.84 & 59.28 & 01.70& 00.86& 82.80 & 72.04 & 53.82 & 38.69 & 76.11 & 64.91 \\
    [2pt]
    SCCDNet-D32  & 65.59  & 49.47 & 03.88& 02.04& 78.08 & 65.47 & 52.86 & 37.81 & 71.78 & 56.86 \\ 
    [2pt]
    DeepCrack & 67.84  & 51.94 & 05.10& 02.67& 76.57 & 63.61 & 48.94 & 34.88 & 70.98 & 56.13 \\
    [2pt]
    CrackFormer        & 71.52  & 56.16 & 35.46& 21.65& 79.61 & 67.89 & 55.18 & 40.03 & 72.22 & 58.06 \\
    [2pt]
    SegDecNet++ & 77.80  & 64.14 & 33.02& 20.12& 81.17 & 69.78 & 54.36 & 39.05 & 80.40 & 67.95 \\ [2pt]
    \midrule
    CrossDiff   &  \textbf{91.34}& \textbf{85.54} & \textbf{55.10}& \textbf{38.34}& \textbf{85.30} & \textbf{76.07} & \textbf{55.41} & \textbf{41.11} & \textbf{85.55} & \textbf{75.98} \\
    \bottomrule
\end{tabular}%
}
\end{center}
\caption{The comparison of CrossDiff with SOTA segmentation methods over the five datasets evaluated by Dice Score and IoU. The best results are highlighted in bold.}
\label{Experiment}
\end{table*}

\begin{table*}[]\large
\renewcommand\arraystretch{1}
\begin{center}
\resizebox{\textwidth}{!}{%
\begin{tabular}{>{\centering\arraybackslash}m{3cm}>{\centering\arraybackslash}m{1cm}>{\centering\arraybackslash}m{1cm}>{\centering\arraybackslash}m{1cm}>{\centering\arraybackslash}m{1cm}>{\centering\arraybackslash}m{1cm}>{\centering\arraybackslash}m{1cm}>{\centering\arraybackslash}m{1cm}>{\centering\arraybackslash}m{1cm}>{\centering\arraybackslash}m{1cm}>{\centering\arraybackslash}m{1cm}>{\centering\arraybackslash}m{1cm}>{\centering\arraybackslash}m{1cm}}
    \toprule
    \multirow{2}{*}{Threshold ($\mathit{\theta}$)} & \multicolumn{2}{c}{0.1} & \multicolumn{2}{c}{0.3} & \multicolumn{2}{c}{0.5} & \multicolumn{2}{c}{0.7} & \multicolumn{2}{c}{0.9} & \multicolumn{2}{c}{0.95} \\ 
\cmidrule(lr){2-3} \cmidrule(lr){4-5} \cmidrule(lr){6-7} \cmidrule(lr){8-9} \cmidrule(lr){10-11} \cmidrule(lr){12-13}
                    & Dice        & IoU        & Dice        & IoU        & Dice        & IoU        & Dice    & IoU        & Dice        & IoU        & Dice        & IoU        \\ [2pt] \midrule

SegDecNet++  & 65.26 & 49.80 & 68.94 & 53.73 & 71.34 & 56.52 & 73.75 & 59.30 & 77.43 & 63.61 & 75.67 & 62.87           \\
[2pt]
CrossDiff    & 81.17 & 70.77 & 81.72 & 71.61 & 81.72 & 71.61 & 81.72 & 71.61 & 81.31 & 71.01 & 76.64 & 64.20           \\ 
   \bottomrule
\end{tabular}%
}
\end{center}
\caption{The comparison of CrossDiff with the second best method on different thresholds over the five datasets. Average is the average Dice score and IoU of five datasets, weighted by sample numbers.}
\label{Threshold}
\end{table*}

\begin{figure*}
    \centering
    \includegraphics[width=0.85\textwidth]{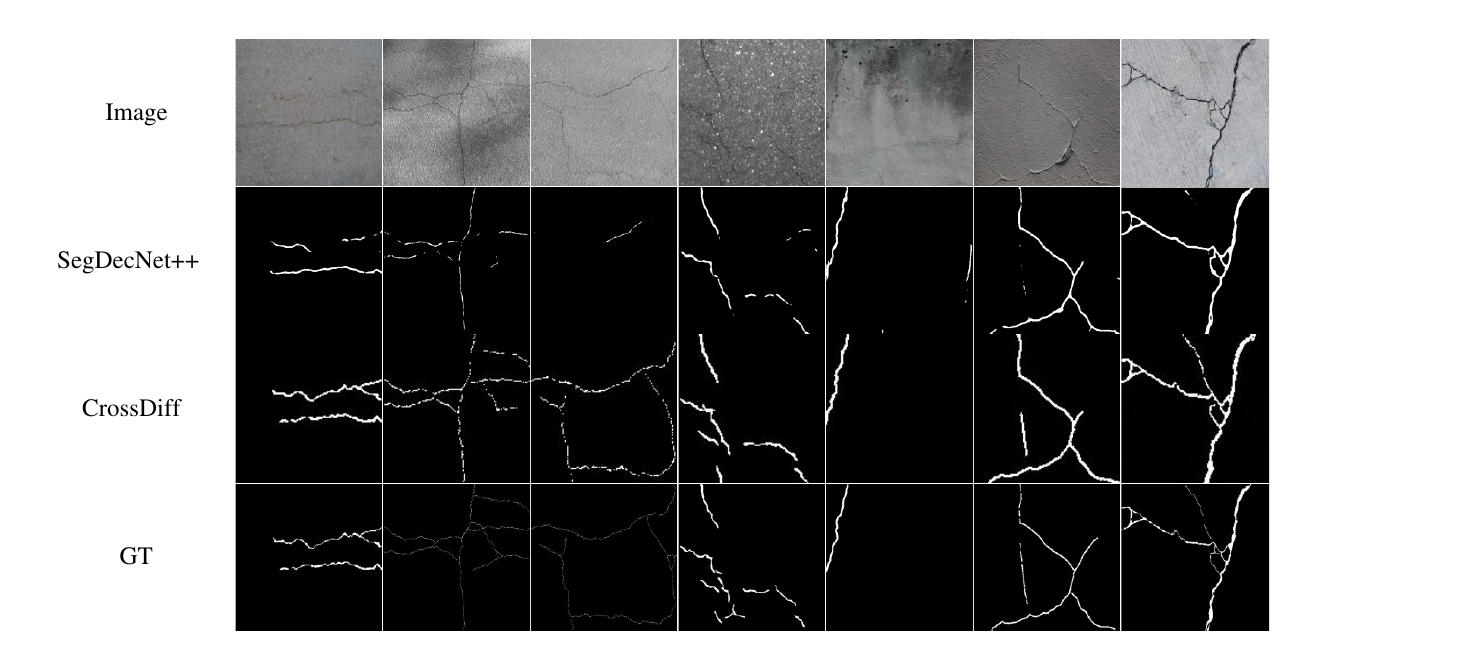}
    \caption{The sequence of images, progressing from top to bottom, includes the input image, SegDecNet++~\cite{tabernik2023automated}, CrossDiff, and ground truth mask (GT). It is apparent from the visual comparison that our model surpasses SegDecNet++~\cite{tabernik2023automated} in terms of the accuracy of cracks localization.}
    \label{fig:show}
\end{figure*}

To verify the slender crack image segmentation performance, we compare CrossDiff with SOTA segmentation methods on multi-crack segmentation datasets. The quantitative results of Dice score and IoU are shown in Table~\ref{Experiment} and Figure~\ref{fig:show} respectively. In Table~\ref{Experiment}, we compare the segmentation methods which are widely used and well-recognized in the community, including the typical image segmentation method DeepLabv3+~\cite{chen2018encoder}(Chen~ 2018), and crack segmentation method the CNN-based method SCCDNet-D32~\cite{li2021sccdnet}(Li~ 2021), DeepCrack~\cite{liu2019deepcrack}(Liu~ 2019), SegDecNet++~\cite{tabernik2023automated}(Tabernik~ 2023) and the transformer-based method CrackFormer~\cite{9711107}(Liu ~ 2021)), in which SegDecNet++ is SOTA method before. 

From Table~\ref{Experiment}, it is evident that CrossDiff outperforms all other methods across five different tasks, underscoring its exceptional generalization prowess across diverse crack segmentation tasks and image modalities. Traditional methods exhibit subpar performance on the CrackTree200 dataset due to the unique characteristics of crack size. In comparison to SegDecNet++, CrossDiff demonstrates an average improvement of 8 percentage points in terms of IoU, especially in CFD and CrackTree200 which feature more slender cracks compared to the other three datasets. Although the IoU achieved by CrossDiff for the CrackTree200 dataset is only 38.34\%, Figure~\ref{fig:show} illustrates that the predictions are capable of roughly localizing the cracks. Furthermore, CrossDiff exhibits superior performance stability at threshold levels compared to the second best method SegDecNet++, as demonstrated in Table~\ref{Threshold}.

\subsection{Ablation Studies}
We do a comprehensive ablation study to verify the effectiveness of the proposed Cross Eecoder and Cross Eecoder. The results are shown in Table \ref{Ablation}, where 448 and 1024 donate the input size of Cross Encoder and the parameter complexity of its ViT backbone (K in Figure~\ref{fig:pipeline}), while layer12 and layer34 donate the number of layers (M in Figure~\ref{fig:pipeline}) in CrossDecoder. We evaluate the performance by average Dice score(\%) and IoU(\%) on all five tasks. Cross Encoder and Cross Decoder improve 1.36\% and 1.35\% in average IoU. However, we observe that as the complexity of the Cross Encoder and Cross Decoder increases, it is difficult for the model to converge.
\begin{table*}[]
\begin{center}
    \resizebox{0.7\textwidth}{!}{%
        \begin{tabular}{>{\centering\arraybackslash}m{2cm}>{\centering\arraybackslash}m{2cm}|>{\centering\arraybackslash}m{2cm}>{\centering\arraybackslash}m{2cm}|>{\centering\arraybackslash}m{2cm}>{\centering\arraybackslash}m{2cm}}
            \toprule
            \multicolumn{2}{c}{Cross Encoder} &  \multicolumn{2}{c|}{Cross Decoder} & \multicolumn{2}{c}{Average} \\
            \cmidrule(lr){1-2}\cmidrule(lr){3-4}\cmidrule(lr){5-6}
                    448       &        1024        &    12 layers  &  34 layers  &    Dice       &    IoU  \\
            \midrule                               
            \usym{2613}       &    \usym{2613}     &   \usym{2613} & \usym{2613} & 79.61  &   68.90 \\  
            \checkmark        &    \usym{2613}     &   \usym{2613} & \usym{2613} & 80.77  &   70.26 \\   
            \rowcolor{gray!20}
            \checkmark        &    \usym{2613}     &   \checkmark  & \usym{2613} & 81.72  &   71.61 \\ 
            \usym{2613}       &    \checkmark      &   \usym{2613} & \checkmark  & 79.85  &   68.84 \\
            \checkmark        &    \usym{2613}     &   \usym{2613} & \checkmark  & 79.74  &   68.74 \\
            \bottomrule
        \end{tabular}%
    }
    \end{center}
    \caption{An ablation study on Cross Encoder and Cross Decoder. Average is the average Dice score and IoU of five datasets, weighted by sample numbers.}
\label{Ablation}
\end{table*}
\section{Conclusion}
In this paper, we introduce a Diffusion Probabilistic Model (DPM)-based approach and demonstrate its effectiveness in segmenting slender cracks. Furthermore, we present a novel cross-conditional DPM architecture, named CrossDiff comprising Cross Encoder and Cross Decoder. Through comparative experiments conducted on five crack datasets, CrossDiff shows superior performance in Dice score and IoU compared to previous state-of-the-art approaches. As the first proposed cross-conditional architecture applying DPM to crack segmentation, we believe that our proposed cross-design will serve as an essential architecture for future research.
\bibliography{aaai25}

\end{document}